\documentclass[10pt,twocolumn,letterpaper]{article}

\usepackage{cvpr}
\usepackage{times}
\usepackage{epsfig}
\usepackage{graphicx}
\usepackage{amsmath}
\usepackage{amssymb}
\usepackage{booktabs}
\usepackage{mathtools}
\usepackage{subfig}
\usepackage{multirow}
\usepackage{diagbox}

\usepackage[pagebackref=true,breaklinks=true,letterpaper=true,colorlinks,bookmarks=false]{hyperref}

\cvprfinalcopy 

\def\cvprPaperID{841} 

\ifcvprfinal\fi
\begin{document}

\title{Peephole: Predicting Network Performance Before Training}

\author{Boyang Deng\thanks{Work done during an internship at SenseTime}\\
Beihang University\\
{\tt\small billydeng@buaa.edu.cn}
\and
Junjie Yan\\
SenseTime\\
{\tt\small yanjunjie@sensetime.com}
\and
Dahua Lin\\
The Chinese University of Hong Kong\\
{\tt\small dhlin@ie.cuhk.edu.hk}
}

\maketitle


\begin{abstract}
\label{sub:abstract}
The quest for performant networks has been a significant force that
drives the advancements of deep learning in recent years. 
While rewarding, improving network design has never been an easy journey.
The large design space combined with the tremendous cost required
for network training poses a major obstacle to this endeavor.
In this work, we propose a new approach to this problem, namely,
predicting the performance of a network before training, based on its architecture.
Specifically, we develop a unified way to encode individual layers
into vectors and bring them together to form an integrated description
via LSTM. Taking advantage of the recurrent network's strong expressive power,
this method can reliably predict the performances of various network architectures.
Our empirical studies showed that it not only achieved accurate predictions but also
produced consistent rankings across datasets -- a key desideratum
in performance prediction.
\end{abstract}


\section{Introduction}
\label{sec:introduction}

The computer vision community has witnessed a series of breakthroughs
over the past several years. What lies behind this remarkable progress
is the advancement in Convolutional Neural Networks (CNNs)~\cite{lecun1998gradient,krizhevsky2012}.
From AlexNet~\cite{krizhevsky2012},
VGG~\cite{simonyan2014},
GoogLeNet~\cite{szegedy2015},
to ResNet~\cite{he2016},
we have come a long way in improving the network design,
which also results in substantial performance improvement.
Take ILSVRC~\cite{deng2009imagenet} for example, the classification error rate
has dropped from $15.3\%$ to below $3\%$ in just a few years, primarily thanks to
the evolution of network architectures.
Nowadays, \emph{``using a better network''} has become a commonly adopted
strategy to boost performance -- this strategy, while simple,
has been repeatedly shown to be very effective in practical applications,
\eg~recognition~\cite{deng2009imagenet}, detection~\cite{lin2014microsoft}, and
segmentation~\cite{everingham2010pascal}.

\begin{figure}[t]
    \begin{center}
        \includegraphics[width=0.86\linewidth]{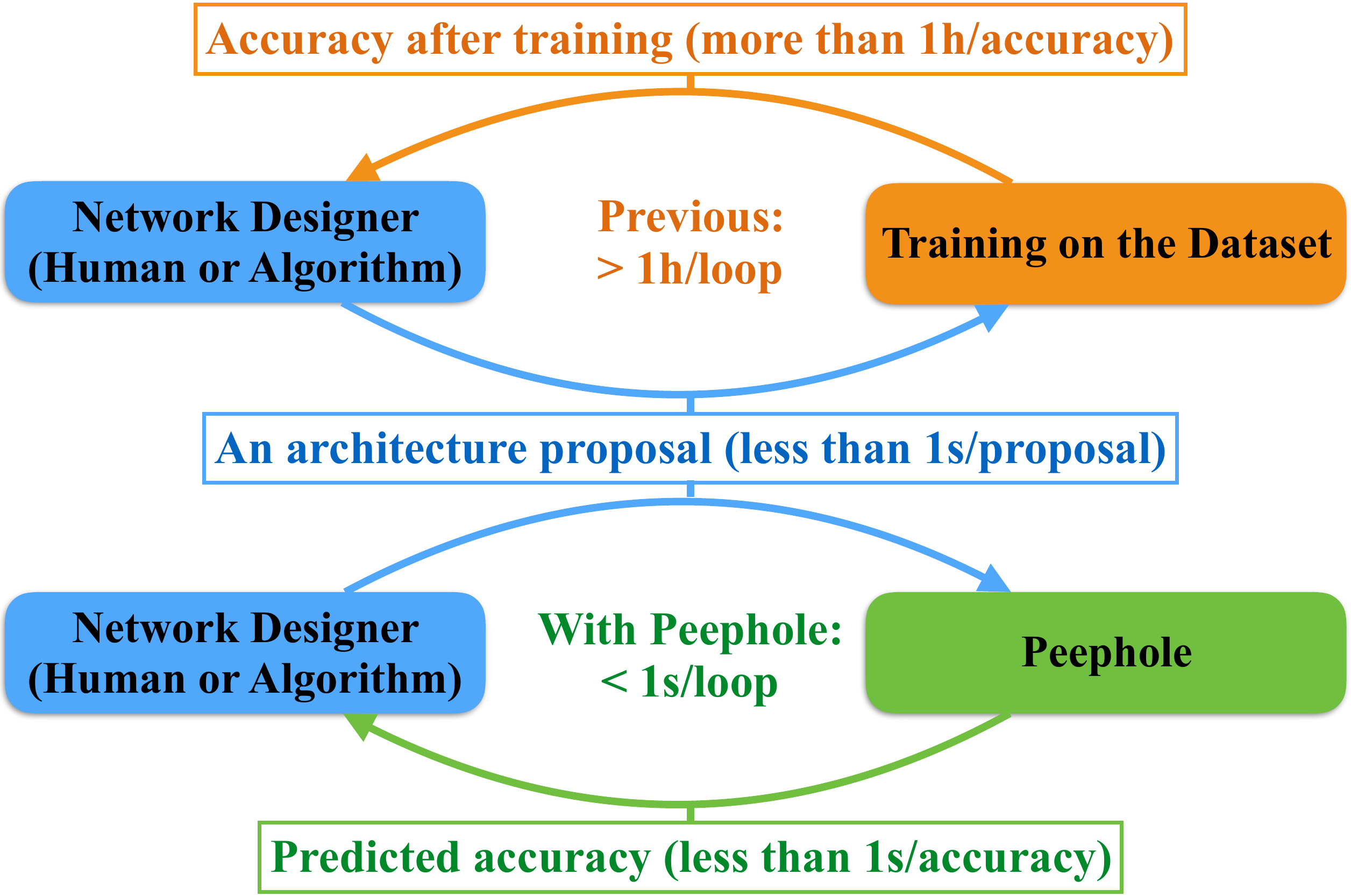}
    \end{center}
    \caption{\small
        \emph{Peephole}, the proposed network performance predictor,
        can dramatically reduce the cost of network design.
        The top circle, which relies on post-training verification,
        takes hours per loop due to the time-consuming training process,
        while the bottom circle, which relies on the proposed predictor,
        can provide fast and effective feedback within $1$ second
        for each architecture.
    }
\label{fig:teaser}
\end{figure}


However, improving network designs is \emph{non-trivial}.
Along this way, we are facing two key challenges, namely,
the \emph{large design space} and the \emph{costly training process}.
Specifically, to devise a convolutional network,
one has to make a number of modeling choices,
\eg~the number of layers, the number of channels
within these layers, and whether to insert a pooling layer
at certain points, etc. All such choices together constitute
a huge design space that is simply beyond our means to conduct
a thorough investigation. Previous efforts were mostly motivated
by intuitions -- though fruitful in early days, this approach has met
increasing difficulties as networks become more complicated.

In recent works~\cite{zoph2016neural,Baker16design,zoph2017learning,zhong2017practical},
automatic search methods have been proposed. These methods
seek better designs (within a \emph{restricted} design space),
by gradually adjusting parts of the networks and validating
the generated designs on real datasets.
Without an effective prior guidance, such search procedures
tend to spend lots of resources in evaluating \emph{``unpromising''}
options.
Also note that training a network in itself is a time-consuming
process. Even on a dataset of moderate size, it may take
hours (if not days) to train a network.
Consequently, an excessively long process is generally needed to
to find a positive adjustment.
It's been reported~\cite{zoph2016neural,zoph2017learning} that
searching a network on CIFAR \emph{takes hundreds of GPUs for a lengthy period}.


Like many others in this community, we have our own share of painful experience
in finding good network designs.
To mitigate this lengthy and costly process,
we develop an approach to quantitatively assess an architecture before
investing resources in training it.
More accurately, we propose a model, called \emph{Peephole},
to predict the final performance of an architecture \emph{before} training.


In this work, we explore a natural idea, that is, to formulate
the network performance predictor as a regression model, which accepts
a network architecture as the input and produces a score as a predictive
estimate of its performance, \eg~the accuracy on the validation set.
Here, the foremost question is
\emph{how to turn a network architecture into a numerical representation}.
This is nontrivial given the diversity of possible architectures.
We tackle this problem in two stages.
First, we develop a vector representation, called \emph{Unified Layer Code},
to encode individual layers.
This scheme allows various layers to be represented
\emph{uniformly} and \emph{effectively} by vectors of a fixed dimension.
Second, we introduce an LSTM network to integrate the layer representations,
which allows architectures with different depths and topologies to be handled
in a uniform way.

Another challenge that we face is \emph{how to obtain a training set}.
Note that this task differs essentially from conventional ones in that
the samples are network architectures together with their performances
instead of typical data samples like images or data records.
Here, the sample space is huge and it is very expensive to obtain even
a sample (which involves running an entire training procedure).
In addressing this issue, we draw inspirations from engineering practice,
and develop a \emph{block-based} sampling scheme,
which generates new architectures by integrating the blocks sampled
from a Markov process.
This allows us to explore a large design space with limited budget while
ensuring that each sample architecture is \emph{reasonable}.


Overall, our main contributions lie in three aspects:
\textbf{(1)} We develop \emph{Peephole},
a new framework for predicting network performance based on
\emph{Unified Layer Code} and \emph{Layer Embedding}.
Our \emph{Peephole} can predict a network's performance
\emph{before} training.
\textbf{(2)} We develop \emph{Block-based Generation}, a simple yet
effective strategy to generate a diverse set of reasonable network architectures.
This allows the proposed performance predictor to be learned with an
\emph{affordable} budget.
\textbf{(3)} We conducted empirical studies over more than a thousand networks,
which show that the proposed framework can make reliable predictions for
a wide range of network architectures and produce consistent ranking
across datasets. Hence, its predictions can provide an effective way to
search better network designs, as shown in Figure~\ref{fig:teaser}.


\section{Related Work}
\label{sec:relwork}

\paragraph{Network Design.}


Since the debut of AlexNet~\cite{krizhevsky2012},
CNNs have become widely adopted to solve computer vision problems.
Over the past several years, the advances in network design have been
a crucial driving force behind the progress in computer vision.
Many representative architectures, such as AlexNet~\cite{krizhevsky2012},
VGGNet~\cite{simonyan2014}, GoogLeNet~\cite{szegedy2015},
ResNet~\cite{he2016}, DenseNet~\cite{huang2016densely}, and DPN~\cite{chen2017dual} are designed
\emph{manually}, based on intuitions and experience.
Nonetheless, this approach has become less rewarding.
The huge design space combined with the costly training procedure makes it
increasingly difficult to obtain an improved design.

Recently, the community has become more interested in an alternative
approach, namely automatic network design.
Several methods~\cite{zoph2016neural,Baker16design,zoph2017learning,zhong2017practical}
have been proposed.
These methods rely on reinforcement learning to learn how to improve a
network design. In order to supervise the learning process,
all these methods rely on actual training processes to provide feedback,
which are very costly, in both time and computational resources.
Our work differs essentially.
Instead of developing an automatic design technique, we focus on a crucial
but often overlooked problem, that is,
\emph{how to quickly get the performance feedback}.

\vspace{-11pt}
\paragraph{Network Performance Prediction.}

As mentioned, our approach is to \emph{predict} network performance.
This is an emerging topic, on which existing works remain limited.
Some previous methods on performance prediction were developed
in the context of hyperparameter optimization, using techniques like
Gaussian Process~\cite{swersky2014freeze} or
Last-Seen-Value heuristics~\cite{li2016hyperband}.
These works mainly focus on designing a
special surrogate function for a better evaluation of hyper-configurations.
There have also been attempts to directly predict network performance.
Most works along this line intend to extrapolate the future part of the
learning curve given the elapsed part.
For this, Domhan \etal~\cite{domhan2015speeding} proposed a mixture of parametric
functions to model the learning curve.
Klein \etal~\cite{klein2016learning} extended this work by replacing
the mixture of functions with a Bayesian Neural Network.
Baker \etal~\cite{baker2017practical} furthered this study by
additionally leveraging the information about network architectures
with hand-crafted features, and using $v$-SVR for curve prediction.

All these works rely on partially observed learning curves to make predictions,
which still involve a partly run training procedure and therefore are time-consuming.
To support large-scale search of network designs, we desire much quicker feedback
and thus explore a fundamentally different but more challenging approach,
that is, to predict the performance purely based on architectures.


\begin{figure*}[t]
    \begin{center}
        \includegraphics[width=0.95\linewidth]{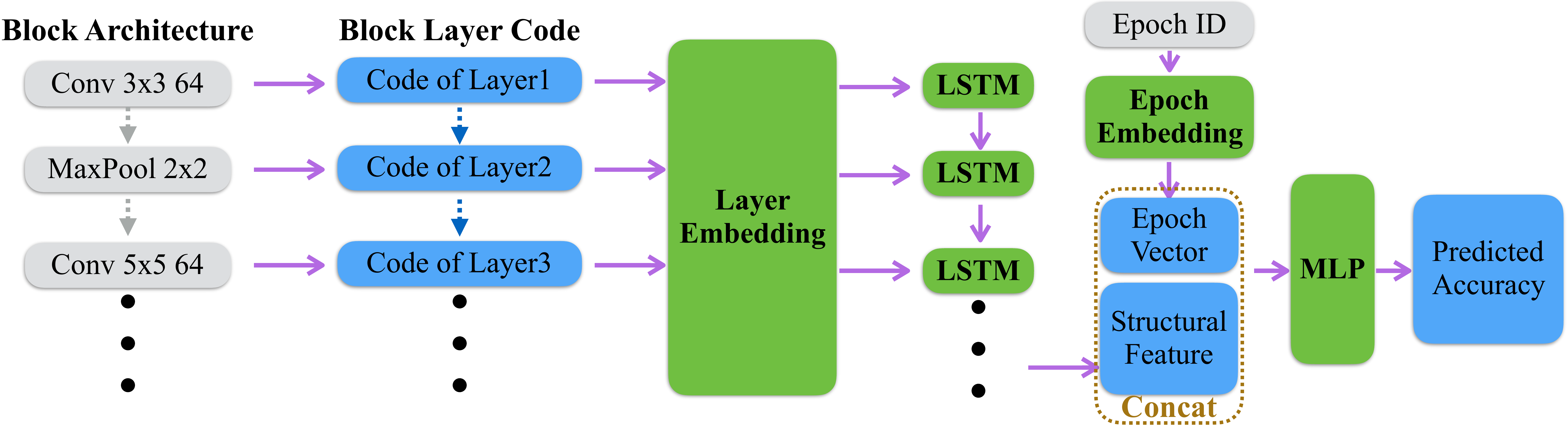}
    \end{center}
    \caption{\small
        The overall pipeline of the \emph{Peephole} framework.
        Given a network architecture, it first encodes each layer into
        a vector through integer coding and layer embedding.
        Subsequently, it applies a recurrent network with LSTM units to
        integrate the information of individual layers following the
        network topology into a \emph{structural feature}.
        This structural feature together with the epoch index (also embedded
        into a vector) will finally be fed to an MLP to predict
        the accuracy at the corresponding time point, \ie~the end of the
        given epoch.
        Note that the blocks indicated by green color,
        including the embeddings, the LSTM, and the MLP,
        are jointly learned in an end-to-end manner.
    }
\label{fig:overview}
\end{figure*}

\section{Network Performance Prediction}
\label{sec:prediction}


Our goal is to develop an effective method to predict network performance
\emph{before} training.
Our predictive model, called \emph{Peephole} and shown in Figure~\ref{fig:overview},
can be formalized as a function, denoted by $f$.
The function $f$ takes two arguments,
a network architecture $x$ and an epoch index $t$, and produces
a scalar value $f(x, t)$ as the prediction of the accuracy
at the end of the $t$-th epoch.
Here, incorporating the epoch index $t$ as an input to $f$ is reasonable,
as the validation accuracy generally changes as the training proceeds.
Therefore, when we predict performance, we have to be specific about
the time point of the prediction.


Note that this formulation differs fundamentally from
previous works~\cite{domhan2015speeding,klein2016learning,baker2017practical}.
Such methods require the observation of the initial part (usually $25\%$)
of the training curve and extrapolate the remaining part.
On the contrary, our method aims to predict the entire curve, relying only
on the network architecture.
In this way, it can provide feedback much quicker and thus is particularly
suited for large-scale search of network designs.


However, developing such a predictor is nontrivial.
Towards this goal, we are facing significant technical challenges,
\eg~\emph{how to unify the representation of various layers},
and \emph{how to integrate the information from individual layers
over various network topologies.}
In what follows, we will present our answers to these questions.
Particularly, Sec.~\ref{subsec:layercode} presents a unified vector representation of layers, 
which is constructed in two steps, namely coding and embedding.
Sec.~\ref{subsec:model} presents an LSTM model for integrating the
information across layers.

\subsection{Unified Layer Code}
\label{subsec:layercode}

\begin{table}[t]\footnotesize
    \begin{center}
        \begin{tabular}{l||cccc}
            \toprule
            Layer & \textbf{TY} & \textbf{KW} & \textbf{KH} & \textbf{CH} \\
            \midrule\midrule
            Conv & $\{1\}$ & $\{1,2,3,4,5\}$ & $\{1,2,3,4,5\}$ &
            $[0.25,3]$ \\
            Max-Pool & $\{2\}$ & $\{2,3,4,5\}$ & $\{2,3,4,5\}$ & $\{1\}$ \\
            Avg-Pool & $\{3\}$ & $\{2,3,4,5\}$ & $\{2,3,4,5\}$ & $\{1\}$ \\
            ReLU & $\{4\}$ & $\{1\}$ & $\{1\}$ & $\{1\}$ \\
            Sigmoid & $\{5\}$ & $\{1\}$ & $\{1\}$ & $\{1\}$ \\
            Tanh & $\{6\}$ & $\{1\}$ & $\{1\}$ & $\{1\}$ \\
            BN & $\{7\}$ & $\{1\}$ & $\{1\}$ & $\{1\}$ \\
            \bottomrule
        \end{tabular}
    \end{center}
    \caption{\small
        The coding table of the Unified Layer Code,
        where each row corresponds to a layer type.
        For each layer, we encode it with
        a \emph{type id (TY)}, a \emph{kernel width (KW)},
        a \emph{kernel height (KH)}, and a \emph{channel number (CH)}.
        Note that for \emph{CH}, we use the ratio of the output
        channel number to the input number, instead of the absolute
        value, and quantize it into an integer.
    }
\label{tab:code}
\end{table}


In general, a convolutional neural network can be considered as
a directed graph whose nodes represent certain operations,
\eg~convolution and pooling. Hence, to develop a representation of
such a graph, the first step is to define a representation
of individual nodes, \ie~the layers.
In this paper, we propose \emph{Unified Layer Code (ULC)},
a uniform scheme to encode various layers into numerical vectors,
which is done in two steps: \emph{integer coding} and \emph{embedding}.


\vspace{-5pt}
\paragraph{Integer coding.}
We notice that the operations commonly used in a CNN, including
convolution, pooling, and nonlinear activation, can all be considered
as applying a kernel to the input feature map. To produce an output value,
the kernel takes a local part of the feature map as input, applies a linear
or nonlinear transform, and then yields an output.
In particular, an element-wise activation function can be considered
as a nonlinear kernel of size $1 \times 1$.

Each operation is also characterized by the number of output channels.
In a typical CNN, the number of channels can vary significantly,
from below $10$ to thousands. However, for a specific layer, this number
is usually decided based on that of the input, according to a ratio within
a limited range.
Particularly, for both pooling and nonlinear activation, the numbers of
output channels are always equal to that of the input channels, and thus
the ratio is $1$. While for convolution, the ratio usually ranges from
$0.25$ to $3$, depending on whether the operation intends to reduce, preserve,
or expand the representation dimension.
In light of this, we choose to represent \emph{CH} by the output-input ratio
instead of the absolute number. In this way, we can effectively limit its
dynamic range, quantize it into $8$ bins
(respectively centered at $0.25, 0.5, 0.75, 1.0, 1.5, 2.0, 2.5, 3.0$).

Overall, we can represent a common operation by a tuple of four integers
in the form of \emph{(TY, KW, KH, CH)}, where
\emph{TY} is an integer id that indicates the type of the computation,
\emph{KW} and \emph{KH} are respectively the width and height of the kernel,
while \emph{CH} represents the ratio of output-input channels (using the
the index of the quantized bin).
The details of this scheme are summarized in Table~\ref{tab:code}.

\label{subsec:embedding}
\begin{figure}[t]
    \begin{center}
        \includegraphics[width=0.95\linewidth]{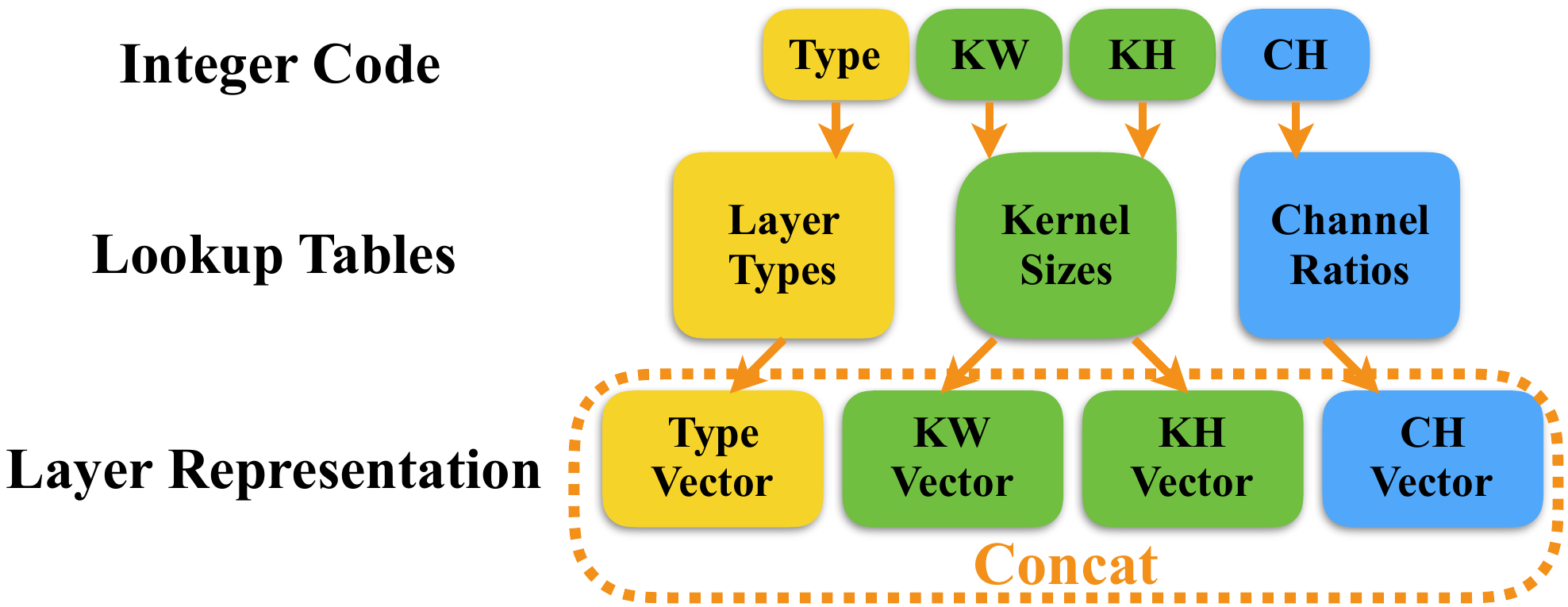}
    \end{center}
    \caption{\small
        The layer embedding component. It takes the integer codes
        as input, maps them to embedded vectors respectively via
        table lookup, and finally concatenates them into a real
        vector representation. Note that \emph{KW} and \emph{KH}
        share the same lookup table.
    }
\label{fig:embedding}
\end{figure}

\vspace{-5pt}
\paragraph{Layer embedding.}
While capturing the key information for a layer,
the \emph{discrete} representation introduced above is not amenable to
complex numerical computation and deep pattern recognition.
Inspired by word embedding~\cite{mikolov2013distributed},
a strategy proven to be very effective in natural language processing,
we take one step further and develop
\emph{Layer Embedding}, a scheme to turn the integer codes into
a unified real-vector representation.

As shown in Figure~\ref{fig:embedding}, the embedding is done by table
lookup. Specifically, this module is associated with three lookup tables,
respectively for \emph{layer types}, \emph{kernel sizes}, and
\emph{channel ratios}. Note that the kernel size table is used to encode
both \emph{KW} and \emph{KH}.
Given a tuple of integers, we can convert its element into a real vector
by retrieving from the corresponding lookup table.
Then by concatenating all the embedded vectors derived respectively
from individual integers, we can form a vector representation of the layer.

\subsection{Integrated Prediction}
\label{subsec:model}

With the layer-wise representations based on Unified Layer Code and Layer
Embedding, the next is to aggregate them into an overall representation
for the entire network.
In this work, we focus on the networks with sequential structures, which
already constitute a significant portion of the networks used in real-world
practice. Here, the challenge is how to cope with varying depths
in a uniform way.

Inspired by the success of recurrent networks in sequential modeling,
\eg~in language modeling~\cite{mikolov2010recurrent} and video analytics~\cite{yue2015beyond},
we choose to explore recurrent networks in our problem.
Specifically, we adopt the \emph{Long-Short Term Memory (LSTM)}~\cite{hochreiter1997long},
an effective variant of RNN, for integrating the information along
a sequence of layers.
In particular, an LSTM network is composed of a series of LSTM units,
each for a time step (\ie~a layer in our context). The LSTM maintains
a hidden state $h_t$ and a cell memory $c_t$,
and uses an input gate $i_t$, an output gate $o_t$, and a forget gate $f_t$
to control the information flow. At each step, it takes an input $x_t$,
decides the value of all the gates, yields an output $u_t$, and updates
both the hidden state $h_t$ and the cell memory $c_t$, as follows:
\begin{align}
    i_t &= \sigma\left(W^{(i)}x_t + U^{(i)}h_{t-1} + b^{(i)}\right), \notag \\
    o_t &= \sigma\left(W^{(o)}x_t + U^{(o)}h_{t-1} + b^{(o)}\right), \notag \\
    f_t &= \sigma\left(W^{(f)}x_t + U^{(f)}h_{t-1} + b^{(f)}\right), \notag \\
    u_t &= \tanh\left(W^{(u)}x_t + U^{(u)}h_{t-1} + b^{(u)}\right), \notag \\
    c_t &= i_t \odot u_t + f_t\odot c_{t-1}, \notag \\
    h_t &= o_t \odot \tanh\left(c_t\right).
    \label{eq:LSTM}
\end{align}
Here, $\sigma$ denotes the sigmoid function while $\odot$ the element-wise
product. Along the way from low-level to high-level layers, the LSTM network
would gradually incorporate layer-wise information into the hidden state.
At the last step, \ie~the layer right before the fully connected layer
for classification, we extract the hidden state of the LSTM cell to represent
the overall structure of the network, which we refer to as the
\emph{structural feature}.

As shown in Figure~\ref{fig:overview}, the \emph{Peephole} framework will
finally combine this structural feature with the epoch index (also embedded
into a real-vector) and use a Multi-Layer Perceptron (MLP) to make the final
prediction of accuracy.
In particular, the MLP component at the final step is comprised of
three fully connected layers with Batch Normalization and ReLU activation.
The output of this component is a real value that serves as an estimate
of the accuracy.


\section{Training Peephole}
\label{sec:training}

Like other predictive models, \emph{Peephole} requires sufficient
training samples to learn its parameters.
However, for our problem, the preparation of the training set itself
is a challenge. Randomly sampling sequences of layers is not a viable
solution for two reasons:
(1) The design space grows exponentially as the number of layers
increases, while it is expensive to obtain a training sample
(which requires running an entire training procedure to obtain a performance
curve). Hence, it is unaffordable to explore the entire design space
freely, even with a large amount of computational resources.
(2) Many combinations of layers are not reasonable options from a practical
point of view (\eg~a network with multiple activation layers stacked
consecutively in a certain part of the layer sequence).
Training such networks are simply a waste of resources.

In this section, we draw inspirations from existing practice and
propose a \emph{Block-based Generation} scheme to acquire training samples
in Sec.~\ref{subsec:block}. Then, we present a learning objective
for supervising the training process in Sec.~\ref{subsec:training}.

\begin{figure}[t]
    \begin{center}
        \includegraphics[width=0.83\linewidth]{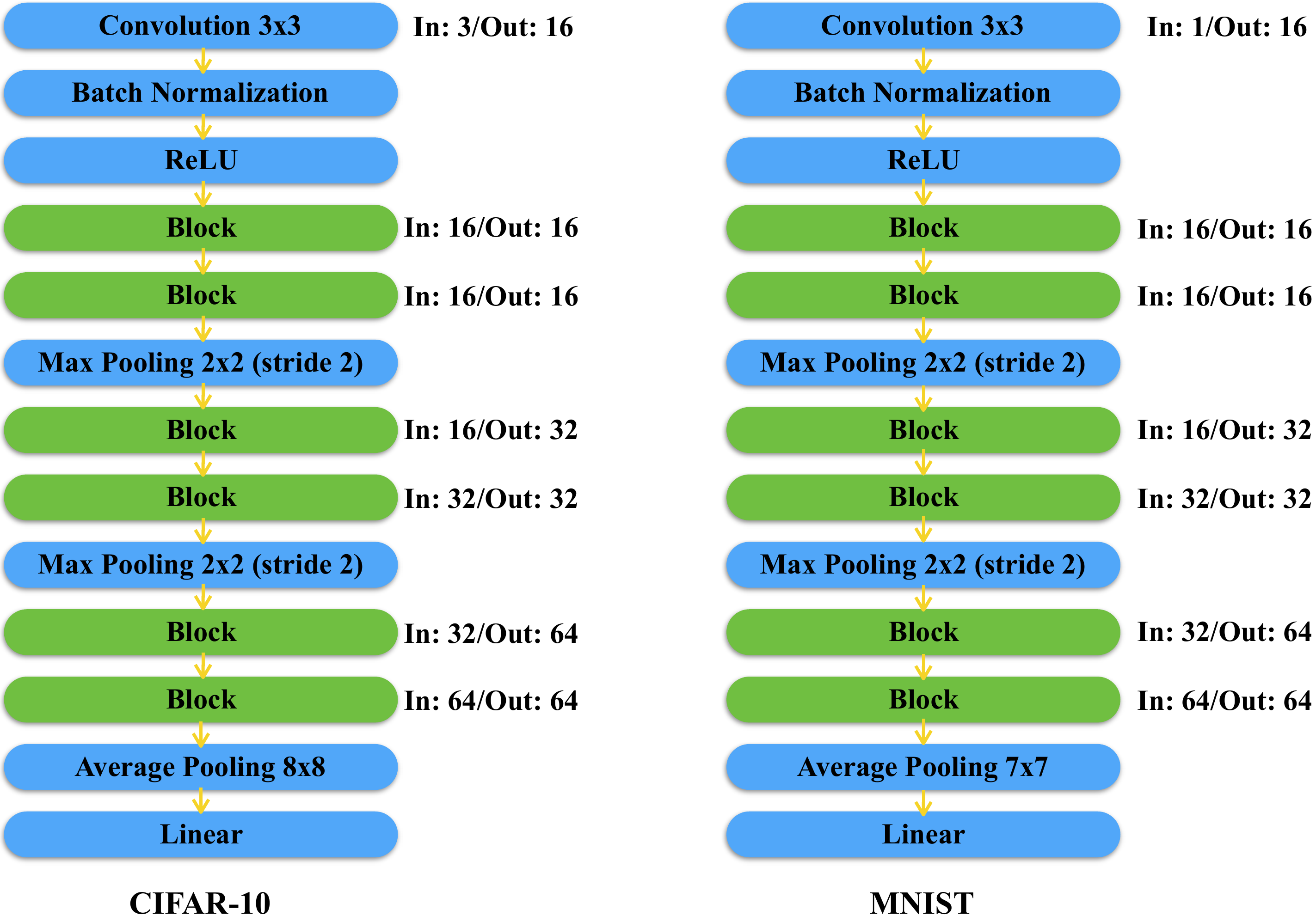}
    \end{center}
    \caption{\small
        Block-based Generation.
        We first generate individual blocks (indicated by green color),
        and then assemble them into a network following a skeleton
        predefined for each dataset (\emph{MNIST} or \emph{CIFAR}).
        The layers in blue color are \emph{fixed} in the skeleton.
        Different networks for a dataset differ mainly in the block
        designs.
    }
\label{fig:stack}
\end{figure}

\subsection{Block-based Generation}
\label{subsec:block}


The engineering practice of network design~\cite{he2016,xie2016aggregated}
suggests that it is a good strategy to construct a neural network by stacking
blocks that are structurally alike.
Zoph \etal~\cite{zoph2017learning} and Zhong \etal~\cite{zhong2017practical}
also proposed to search transferable blocks
(referred to as \emph{cells} in~\cite{zoph2017learning})
and assemble them into a network in their efforts towards an automatic
way for network search.
Inspired by these works, we propose \emph{Block-based Generation},
a simple yet effective strategy to prepare our training samples.
As illustrated in Figure~\ref{fig:embedding}, 
it first designs individual blocks and then
stacks them into a network following a certain skeleton.


A block is defined to be a short sequence of layers with no more than
$10$ layers.
To generate a block, we follow a Markov chain.
Specifically, we begin with a convolution layer by randomly choosing
its kernel size from $\{1, \ldots, 5\}$ and
the ratio of output/input channel numbers from
$\{0.25, 0.50, 0.75, 1.0, 1.5, 2.0, 2.5, 3.0\}$.
Then at each step, we draw the next layer conditioned on the current one
following predefined transition probabilities between layer types,
which are empirically estimated from practical networks.
For example, a convolution layer has a high chance to be followed by
a batch normalization layer and a nonlinear activation.
An activation layer is more likely to ensued by another convolution
layer or a pooling layer.
More details will be provided in the supplemental materials.


With a collection of blocks, we can then build
a complete network by assembling them following a skeleton.
The design of the skeleton follows the general practice in
computer vision. As shown in Figure~\ref{fig:stack}, the skeleton
comprises three stages with different resolutions. Each stage
is a stack of blocks followed by a max pooling layer to reduce
the spatial resolution. The features from the last block will
go through an average pooling layer and then a linear layer
for classification.
When replicating blocks within a stage, $1 \times 1$ convolution
layers will be inserted in between for dimension adaptation when
the output dimension of the preceding layer does not match
the input dimension of the next layer.


The \emph{block-based generation} scheme presented above effectively
constrain the sample space, ensuring that the generated networks are mostly
reasonable and making it feasible to prepare a training set with an affordable
budget.

\subsection{Learning Objective}
\label{subsec:training}

Given a set of sample networks $\{x_i\}_{1:N}$,
we can obtain a performance curves $y_i(t)$ for each network $x_i$,
\ie~the validation accuracy as a function of epoch numbers,
by training the network on a given dataset.
Hence, we can obtain a set of pairs $\mathcal{D} = \{(x_i, y_i)\}_{1:N}$
and learn the parameters of the predictor in a supervised way.

Specifically, we formulate the learning objective with the \emph{smooth L1} loss,
denoted by $l$, as below:
\begin{equation}
    \mathcal{L}(\mathcal{D}; \boldsymbol\theta) =
    \frac{1}{N} \sum_{i=1}^n l \left(f(x_i, T), y_i(T)\right).
\end{equation}
Here, $\boldsymbol\theta$ denotes the predictor parameters.
Note that we train each sample network with $T$ epochs, and use the results of
the final epoch to supervise the learning process.
Our framework is very flexible -- with the entire learning curves, in principle,
one can use the results at multiple epochs for training.
However, we found empirically that using only the final epochs already yields
reasonably good results.


\subsection{Discussions}

First, the \emph{Peephole} predictor is task-specific. It is trained to
predict the performance on a certain dataset with a specific performance metric.
Second, besides network architectures and epoch numbers,
the performance of a network also depends on a number of other factors,
\eg~how it is initialized, how the learning rate is adjusted over time,
as well as the settings on the optimizers.
In this work, we train all sample networks with a fixed set of or such
design choices.

Admittedly, such a setting may sound a bit restrictive.
However, this actually reflects our typical practice when tuning network
designs in ablation studies. Moreover, most automatic network search schemes
also fix such choices during the search process in order to fairly
compare among architectures.
Therefore, the predictor trained in this way can already provide good support
to the practice.
That being said, we do plan to incorporate additional factors in the predictor
in our future exploration.


\section{Experiments}
\label{sec:experiments}

We tested \emph{Peephole}, the proposed network performance prediction
framework on two public datasets,
CIFAR-10~\cite{krizhevsky2009cifar} and MNIST~\cite{lecun1998mnist}.
Sec.~\ref{subsec:configuration} presents the experiment settings, including
how the datasets are used and the implementation details of our framework.
Sec.~\ref{subsec:results} presents the results we obtained on
both datasets, and compares them with other performance prediction methods.
Sec.~\ref{subsec:imagenet} presents preliminary results on using \emph{Peephole}
to guide the search of better networks on ImageNet.
Finally, Sec.~\ref{subsec:studies} presents a qualitative study on the
learned representations via visualization.

\begin{table}[t]
    \begin{center}
        \begin{tabular}{l||cc|cc}
            \toprule
            & \multicolumn{2}{c|}{\textbf{CIFAR-10}} & \multicolumn{2}{c}{\textbf{MNIST}} \\
            & Train & Validation & Train & Validation \\
            \hline\hline
            Size & $844$ & $800$ & $471$ & $400$ \\
            \hline
            Mean & $0.74$ & $0.73$ & $0.98$ & $0.98$ \\
            \hline
            Std & $0.07$ & $0.08$ & $0.03$ & $0.04$ \\
            \hline
            Min & $0.50$ & $0.48$ & $0.52$ & $0.49$ \\
            \hline
            Max & $0.89$ & $0.88$ & $0.99$ & $0.99$ \\
            \bottomrule
        \end{tabular}
    \end{center}
    \caption{\small
        The performance statistics for the networks respectively sampled
        for CIFAR-10 and MNIST. Here, \emph{Size} means the number of
        sampled networks, \emph{Mean}, \emph{Std}, \emph{Min}, and \emph{Max}
        are respectively the mean, standard deviation, minimum, and maximum
        of the validation accuracies over the corresponding sets of
        networks.
    }
\label{tab:stats}
\end{table}

\subsection{Experiment Configurations}
\label{subsec:configuration}

\paragraph{Datasets.}
CIFAR-10~\cite{krizhevsky2009cifar} is a dataset for object classification.
In recent years, it usually serves as the testbed for convolutional network
designs.
MNIST~\cite{lecun1998mnist} is a dataset for hand-written digit classification,
one of the early and widely used datasets for neural network research.
Both datasets are of moderate scale. We chose them as the basis for our study
because it is affordable for us to train over a thousand networks thereon
to investigate the effectiveness of the proposed predictor.
After all, our goal is to explore an performance prediction method that works
with diverse architectures instead of pursing a state-of-the-art network
on large-scale vision benchmarks.
To prepare the samples for training and validation, we follow the procedure
described in Sec.~\ref{sec:training} to generate two sets of networks,
respectively for CIFAR-10 and MNIST, and train them to obtain performance
curves.

\vspace{-5pt}
\paragraph{Detailed settings.}
For fair comparison, we train all sampled networks with the same setting:
We use SGD with momentum $0.9$ and weight decay $0.0001$.
Each epoch loops over the entire training set in random order.
The learning rate is initialized to $0.1$ and scaled down by a factor of $0.1$
every $60$ epochs (for CIFAR-10) or $80$ epochs (for MNIST).
The network weights are all initialized following the scheme in~\cite{he2016}.
Table~\ref{tab:stats} shows the statistics of these networks and their performances.

For the \emph{Peephole} model, we use $40$-dimensional vectors for both layer
embedding and epoch embedding. The dimension of the hidden states in LSTM
is set to $160$. The Multi-Layer Perceptron (MLP) for final prediction
comprises $3$ linear layers, each with $200$ hidden units.

\subsection{Comparison of Prediction Accuracies}
\label{subsec:results}

\paragraph{Methods to compare.}
We compare our \emph{Peephole} method with two representative methods in
recent works:
\begin{enumerate}
\item \emph{Bayesian Neural Network (BNN)}~\cite{klein2016learning}.
This method is devised to extrapolate learning curves given their initial
portions (usually $25\%$ of the entire ones). It represents each curve as
a linear combination of basis functions and uses Bayesian Neural Network
to yield probabilistic extrapolations.

\item \emph{$\nu$-Support Vector Regression ($\nu$-SVR)}~\cite{baker2017practical}.
This method relies on a regression model, $\nu$-SVR, to make predictions.
To predict the performance of a network
this model takes as input both the initial portion of the learning curve
and simple heuristic features derived based on the network architecture.
This method represents the state of the art on this task.
\end{enumerate}

Note that both methods above require the initial portions of the learning
curves while ours can give feedback \emph{purely} based on the network
architecture \emph{before} training.

\vspace{-5pt}
\paragraph{Evalution criteria.}
We evaluate the predictor performances using three criteria:
\begin{enumerate}
\item \emph{Mean Square Error (MSE)}, which directly measures the
deviation of the predictions from the actual values.

\item \emph{Kendall's Tau (Tau)}, which measures the correlation between the
predictive rankings among all testing networks and their actual rankings.
The value of Kendall's Tau ranges from $-1$ to $1$, and a higher value
indicates higher correlation.

\item \emph{Coefficient of Determination ($R^2$)}, which measures how closely
the predicted value depends on the actual value. The value of $R^2$ ranges
from $0$ to $1$, where a value closer to $1$ suggests that the prediction
is more closely coupled with the actual value.
\end{enumerate}

\begin{table}[t]
    \begin{center}
        \begin{tabular}{l|ccc}
            \toprule
            \textbf{Method} & \textbf{MSE} & \textbf{Tau} & \boldmath $R^2$ \\
            \midrule\midrule
            \textbf{BNN} & $0.0032$ & $0.5417$ & $0.5400$ \\
            \textbf{$\nu$-SVR} & $0.0018$ & $0.6232$ & $0.7321$ \\
            \midrule
            \textbf{Peephole (Ours)} & \boldmath $0.0010$ & \boldmath $0.7696$ & \boldmath $0.8596$ \\
            \bottomrule
        \end{tabular}
    \end{center}
    \caption{\small
        Comparison of prediction accuracies for the networks trained on
        CIFAR-10, with three metrics \emph{MSE}, \emph{Tau}, and $R^2$.
        The best result for each metric is highlighted with bold font.
    }
\label{tab:cifar10}
\end{table}

\begin{table}[t]\footnotesize
    \begin{center}
        \begin{tabular}{l|ccc}
            \toprule
            \textbf{Method} & \textbf{MSE} & \textbf{Tau} & \boldmath $R^2$ \\
            \midrule\midrule
            \textbf{BNN} & $0.0004$ & $0.3380$ & $0.1921$ \\
            \textbf{$\nu$-SVR} & $0.0008$ & $0.3489$ & $0.2121$ \\
            \midrule
            \textbf{Peephole (Ours)} & \boldmath $0.0003$ & \boldmath $0.5036$ & \boldmath $0.4681$ \\
            \bottomrule
        \end{tabular}
    \end{center}
    \caption{\small
        Comparison of prediction accuracies for the networks trained on
        CIFAR-10, with three metrics \emph{MSE}, \emph{Tau}, and $R^2$.
        The best result for each metric is highlighted with bold font.
    }
\label{tab:mnist}
\end{table}

\vspace{-10pt}
\paragraph{Results on CIFAR-10.}
Table~\ref{tab:cifar10} compares the prediction results for the networks
trained on CIFAR-10, obtained with different predictors. We observe that
\emph{Peephole} consistently outperforms both \emph{BNN} and \emph{$\nu$-SVR}
across all metrics.
Particularly, achieving smaller MSE means that the predictions from
\emph{Peephole} are generally more accurate than those from others.
This makes it a viable predictor in practice.
On the other hand, the high values in \emph{Tau} and $R^2$ indicate that
the ranking among multiple networks produced by \emph{Peephole}
is quite consistent with the ranking of their actual performances.
This makes \emph{Peephole} a good criterion to select performant
network architectures.

\begin{figure}[t]
    \begin{center}
        \subfloat[BNN]{\includegraphics[width=0.32\linewidth]{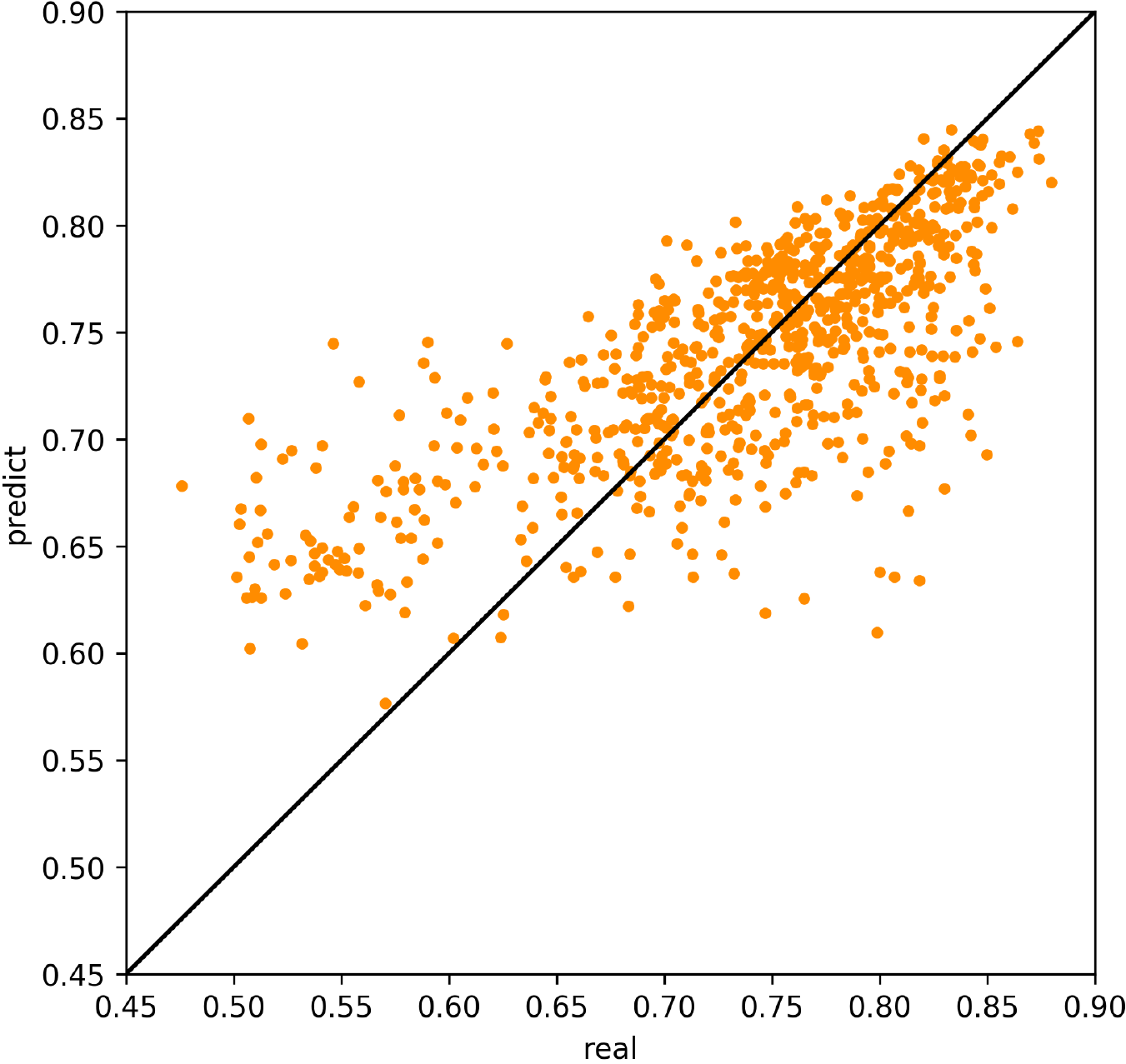}\label{fig:bnn}}
        \hspace{0.06cm}
        \subfloat[$v$-SVR]{\includegraphics[width=0.32\linewidth]{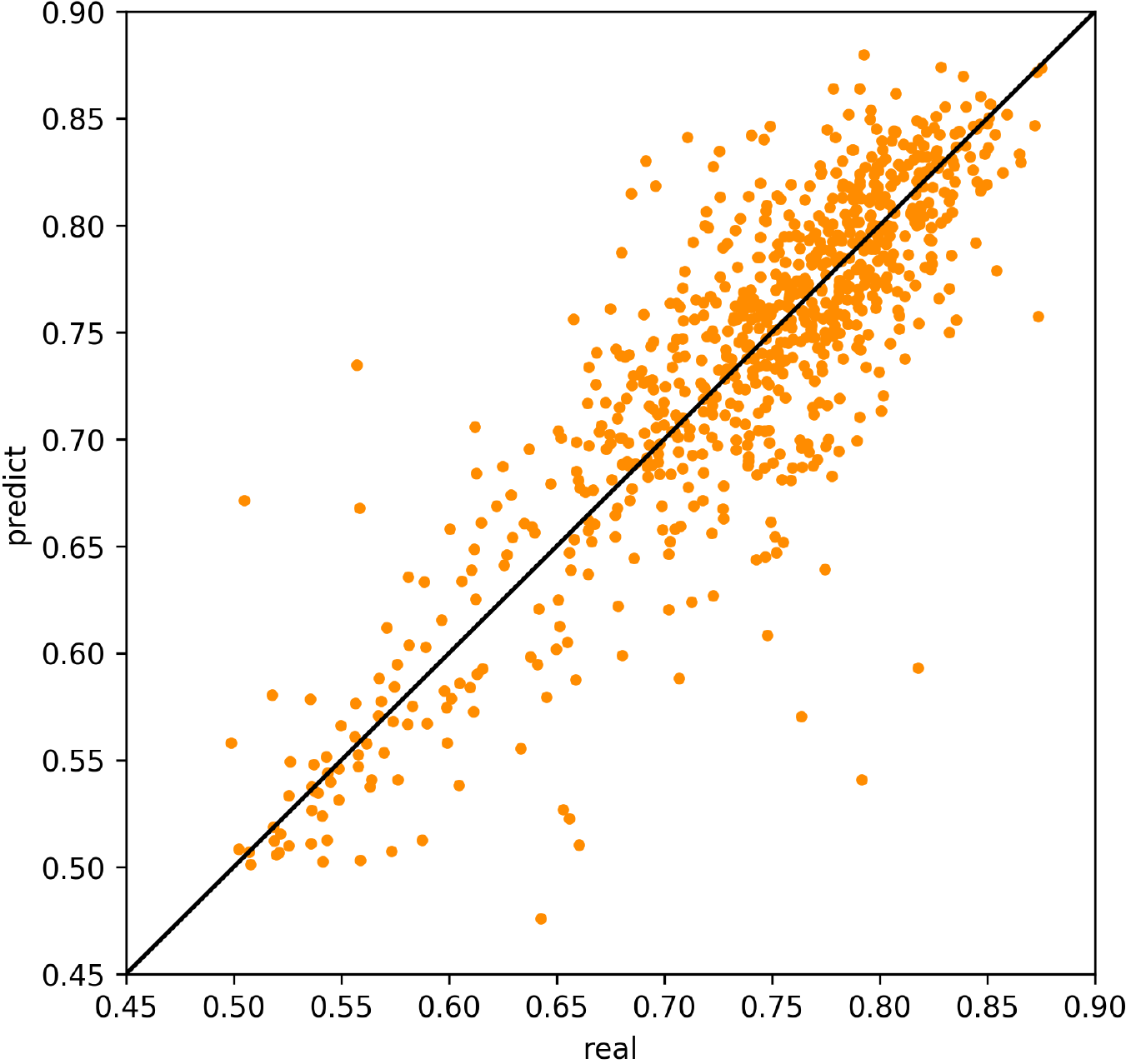}\label{fig:svr}}
        \hspace{0.06cm}
        \subfloat[Peephole]{\includegraphics[width=0.32\linewidth]{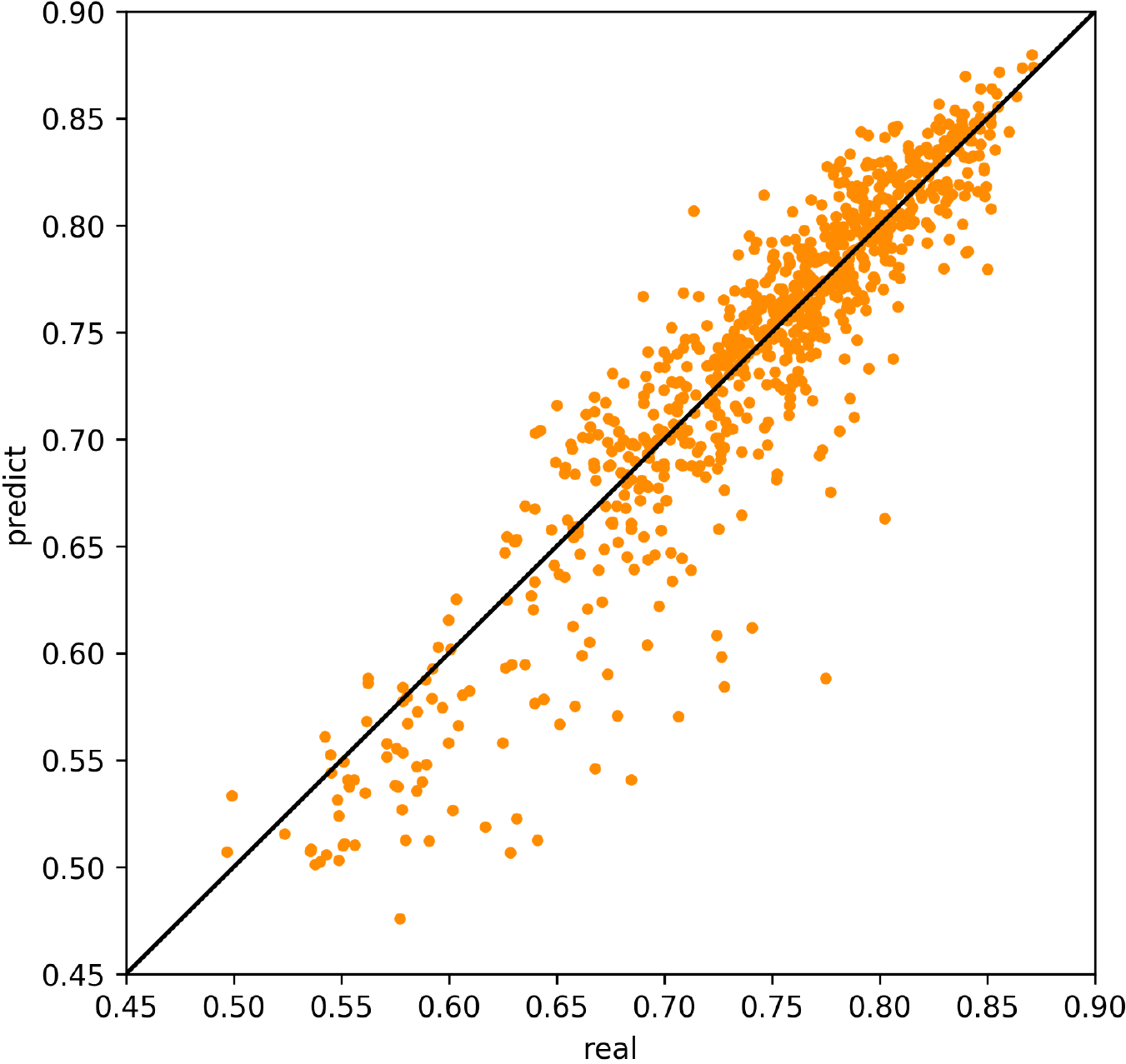}\label{fig:peephole}}
    \end{center}
    \caption{\small
        The scatter plots with x-axes representing the actual validation
        accuracies on CIFAR-10 and y-axes representing the predicted
        accuracies. These plots show how much the predictions are correlated
        with the actual values.
    }
\label{fig:scatter}
\end{figure}

The scatter plots in Figure~\ref{fig:scatter} visualize the correlations
between the predicted accuracies and actual accuracies, obtained with
different methods. Qualitatively, the predictions made by \emph{Peephole}
demonstrate notably higher correlation with the actual values than those from
other methods, especially at the high-accuracy area (top right corner).

\vspace{-10pt}
\paragraph{Results on MNIST.}
We also evaluated the predictions on the networks trained on MNIST
in the same way, with the results shown in Table~\ref{tab:mnist}.
Note that since most networks can yield high accuracies on this MNIST,
it would be easier to produce more precise predictions on the accuracy
numbers but more difficult to yield consistent rankings.
This is reflected by the performance metrics in the table.
Despite this difference in data characteristics,
\emph{Peephole} still significantly outperforms the other two methods
across all metrics.

\begin{table}[t]\footnotesize
    \begin{center}
        \begin{tabular}{l|c|cc|cc}
            \toprule
            \multirow{2}{*}{\textbf{Model}} & \multirow{2}{*}{\textbf{\#Param}} &
            \multicolumn{2}{c|}{\textbf{CIFAR-10}} & \multicolumn{2}{c}{\textbf{ImageNet}}
            \\
            & & Pred & Real & Top-1 & Top5 \\
            \midrule\midrule
            \textbf{VGG-13} & $126$M & -- & -- & $71.55\%$ & $90.37\%$ \\
            \textbf{Selected} & $35$M & $87.52\%$ & $88.06\%$ & $71.91\%$ & $90.53\%$ \\
            \bottomrule
        \end{tabular}
    \end{center}
    \caption{\small
        The accuracies on ImageNet obtained by a network selected based on
        \emph{Peephole} predictions. Here,
        \emph{Pred} is the predicted accuracy on CIFAR-10.
        \emph{Real} is the actual accuracy on CIFAR-10.
        \emph{\#Param} is the number of parameters.
        \emph{Top-1} and \emph{Top-5} indicates the accuracies obtained
        on ImageNet.
    }
\label{tab:imagenet}
\end{table}

\subsection{Transfer to ImageNet}
\label{subsec:imagenet}

Getting top performance on ImageNet is a holy grail for convolutional
network design.
Yet, directly training \emph{Peephole} based on ImageNet is prohibitively
expensively due to the lengthy processes of training a network on ImageNet.
Nevertheless, \cite{zoph2017learning} suggests an alternative way,
that is, to search for scalable and transferable block architectures
on a smaller dataset like CIFAR-10.
Following this idea, we select the network architecture with the
highest \emph{Peephole-predicted accuracy} among those in our validation set
for CIFAR-10, then scale it up and transfer it to ImageNet\footnote{The details
of the selected network and this transfer process will be provided in the
supplemental materials.}.

We compared this network with VGG-13~\cite{simonyan2014},
a widely used network that was designed manually.
From the results in Table~\ref{tab:imagenet}, we can see that
the selected network achieves moderately better accuracy
on ImageNet with a substantially smaller parameter size.
This is just a preliminary study. But it shows that \emph{Peephole} is promising
for pursuing performant network designs that are transferable to the
larger datasets.

\subsection{Studies on the Representations}
\label{subsec:studies}

\begin{figure}[t]
    \begin{center}
        \includegraphics[width=0.95\linewidth]{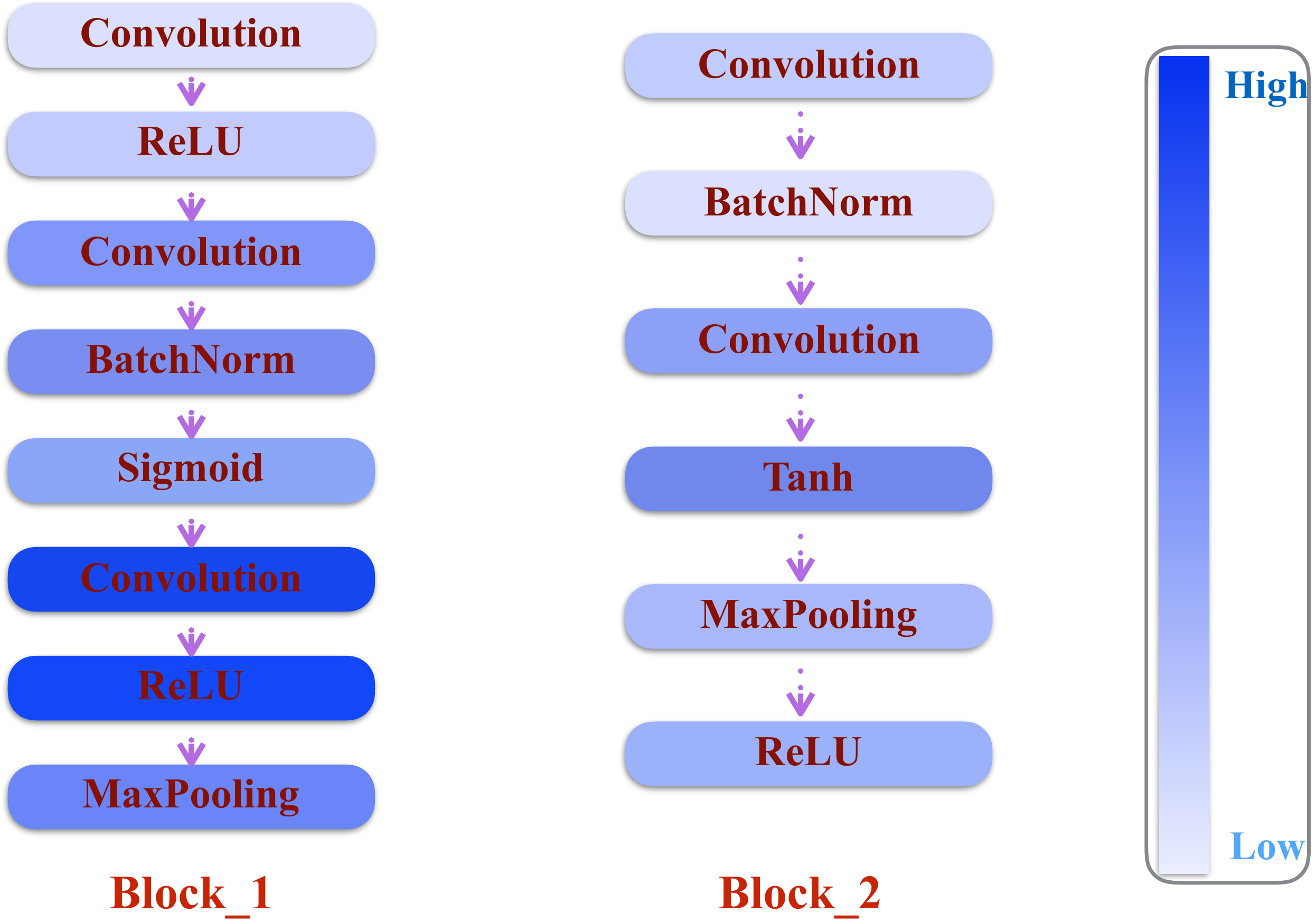}
    \end{center}
    \caption{\small
        Responses of a cell in the learned LSTM for two block architectures. The
        color is drawn according to the response value of this cell. On the right is the
        color bar used in this figure, where darker color represents higher responses. Every time a convolution
        layer appears the response will get higher (the color gets darker in the figure). This
        figure is better viewed in color.
    }
\label{fig:lstm}
\end{figure}

The effective of \emph{Peephole} may be attributed to its ability to abstract
a uniform but expressive representation for various architectures.
To gain better understanding of this representation, we analyze
the learned LSTM and the derived feature vectors.

In one study, we examined the hidden cells inside the LSTM using the
method presented in~\cite{karpathy2015visualizing}.
Particularly, we recorded the dynamics of the cell responses as the LSTM
traverses the sequence of layers. Figure~\ref{fig:lstm} shows the responses
of a certain cell, where we can see that the response raises every time
it gets to a convolution layer. This behavior is observed in different blocks.
This observation suggests that this cell learns to \emph{``detect''}
convolution layers even without being explicitly directed to do.
In a certain sense, this also reflects the capability of LSTM to capture
architectural patterns.

\begin{figure}[t]
    \begin{center}
        \includegraphics[width=0.88\linewidth]{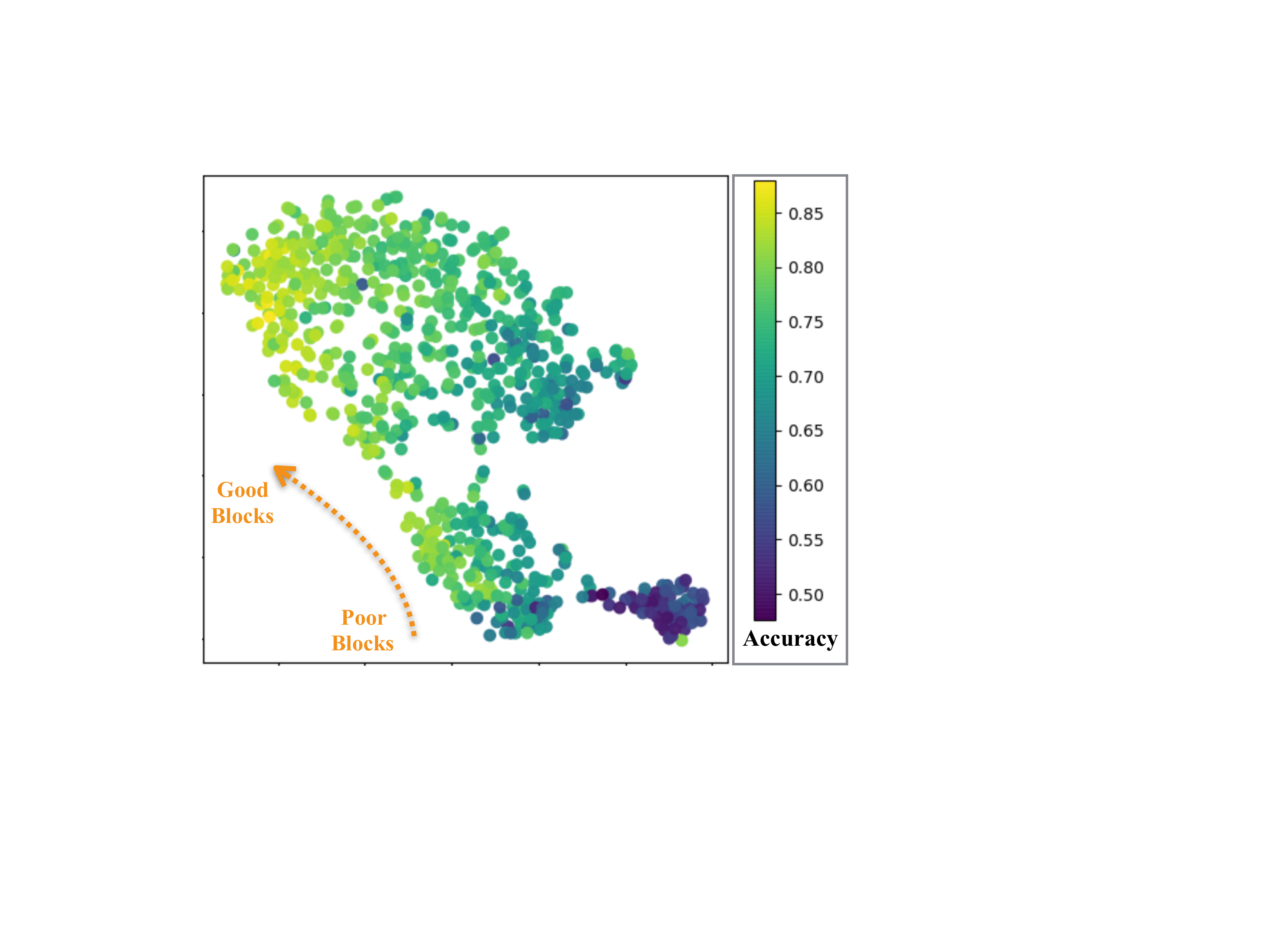}
    \end{center}
    \caption{\small
        The learned structural features for the networks in the validation set,
        visualized based on t-SNE embeddings.
        Points are colored according to their accuracies.
        (better viewed in color)}
\label{fig:tsne}
\end{figure}

In another study, we visualized the structural feature (derived from the
last unit of the LSTM) using t-SNE embedding~\cite{maaten2008visualizing}.
Figure~\ref{fig:tsne} shows the visualized results.
We can see the gradual transition from low performance networks to
high performance networks. This shows that the structural features
contain key information related to the network performances.


\section{Conclusion}
\label{sec:conclusion}

We presented \emph{Peephole}, a predictive model for
predicting network performance based on architectures \emph{before}
training.
Specifically, we developed \emph{Unified Layer Code} as a unified
representation for network architectures and a LSTM-based model to
integrate the information from individual layers.
To tackle the difficulties in preparing the training set,
we propose a \emph{Block-based Generation} scheme, which allows us to
explore a wide variety of reasonable designs while constraining
the search space.
The systematic studies with over a thousand networks trained on CIFAR-10
and MNIST showed that the proposed method can yield reliable predictions
that are highly correlated with the actual performance.
On three different metrics, our method significantly outperforms
previous methods.

We note that this is just the first step towards the goal of fast search
of network designs. In future work, we plan to incorporate additional
factors in our predictor, such as various design choices, to extend
the applicability of the predictive model. We will also explore
more effective ways to optimize network designs on top of this predictive model.

{\small
\bibliographystyle{ieee}
\bibliography{reference}
}


\newpage
\appendix
\setcounter{table}{0}
\setcounter{figure}{0}
\section{Appendix}
\paragraph{Selected Block for ImageNet.}

\begin{figure}[t]
    \begin{center}
        \includegraphics[width=0.95\linewidth]{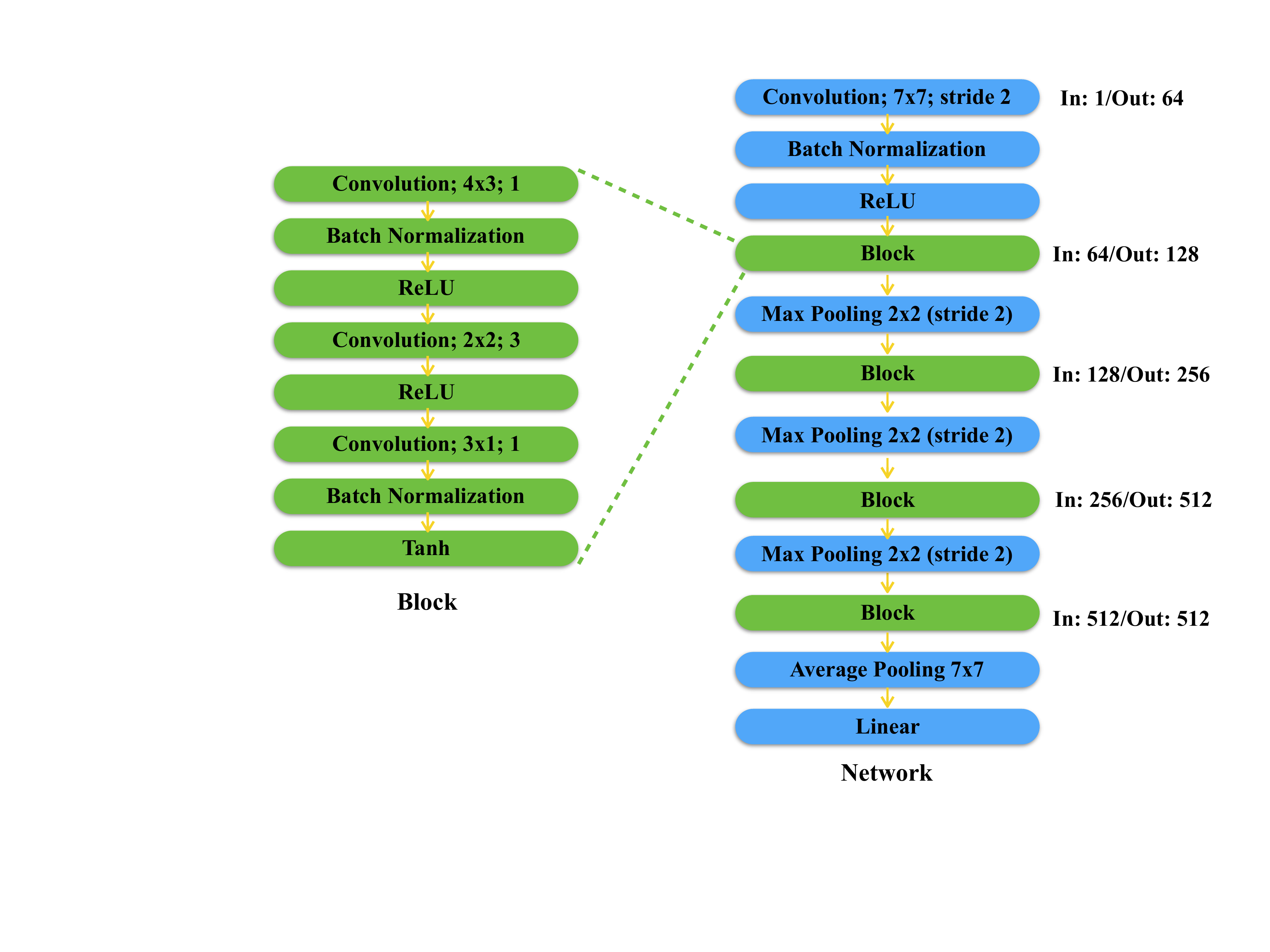}
    \end{center}
    \caption{Selected Architecture for ImageNet. On the left is the block
        architecture selected based on \emph{Peephole}-predicted accuracy. We mark the
    type of each layer. For Convolution layers, we also note the kernel size
($Width\times Height$) and the ratio of output/input channel sizes (the last digit). On
the right is the scheme we use to stack replicants of the left block into a network, which
achieves $71.91\%$ Top-1 accuracy.}
\label{fig:selected}
\end{figure}

In Figure~\ref{fig:selected}, we illustrate the selected block architecture based on
\emph{Peephole}-predicted accuracy. 
We also stack them in a similar manner to our scheme used in
CIFAR-10. 
Note that this architecture is not generated by our algorithm but selected from 
randomly sampled validation architectures using \emph{Peephole}.

\begin{table}[b]\footnotesize
    \begin{center}
        \begin{tabular}{c||c|c|c|c|c|c}
            \hline
            \diagbox{Now}{Next} & \textbf{Conv} & \textbf{MP} & \textbf{AP} & \textbf{ReLU} & \textbf{Sigm} &
            \textbf{Tanh} \\
            \hline\hline
            \textbf{Conv} & $0.03$ & $0.03$ & $0.03$ & $0.31$ & $0.3$ & $0.3$ \\
            \hline
            \textbf{MP} & $0.6$ & $0.05$ & $0.05$ & $0.1$ & $0.1$ & $0.1$ \\
            \hline
            \textbf{AP} & $0.6$ & $0.05$ & $0.05$ & $0.1$ & $0.1$ & $0.1$ \\
            \hline
            \textbf{ReLU} & $0.3$ & $0.3$ & $0.3$ & $0$ & $0.05$ & $0.05$ \\
            \hline
            \textbf{Sigm} & $0.3$ & $0.3$ & $0.3$ & $0.05$ & $0$ & $0.05$ \\
            \hline
            \textbf{Tanh} & $0.3$ & $0.3$ & $0.3$ & $0.05$ & $0.05$ & $0$ \\
            \hline
        \end{tabular}
    \end{center}
    \caption{Transition Matrix of the Markov Chain in Block-based Generation. Every row is the
    distribution of the next layer conditioned on the current layer (denoted as Now in the Table). 
    Conv stands for Convolution.
    MP refers to Max-Pooling.
    AP refers to Average-Pooling.
    Sigm represents Sigmoid.
    Note that Batch Normalization is not illustrated in this table since we insert them right behind the
Convlution layers with the probability of $0.6$.}
\label{tab:transition}
\end{table}

\paragraph{Details of Sampling Strategy.}

Here we detail our configurations for Block-based Generation scheme.
The whole process begins with a convolution layer whose kernel size is uniformly sampled
from $\{1, \dots, 5\}$ and the ratio of output/input channel number is uniformly sampled
from $\{0.25, 0.50, 0.75, 1.0, 1.5, 2.0, 2.5, 3.0\}$.
Then the construction will follow a Markov chain, \ie~we will choose the type of the next
layer merely based on the current one. The transition matrix is shown in
Table~\ref{tab:transition}.
Note that Batch Normalization is inserted right behind Convolution layers with the
probability of $0.6$. Thus it's not shown in the table.
Meanwhile, for computational consideration, we limit the depth of a block to less than
$11$ layers and restrict the number of convolution layers within a block to less than $4$.

\end{document}